\documentclass{article}
\usepackage{spconf,amsmath,graphicx,hyperref}
\usepackage{booktabs}
\usepackage{float}

\title{Cross-Lingual Multi-Granularity Framework for Interpretable Parkinson's Disease Diagnosis from Speech}
\threeauthors
 {Ilias Tougui}
	{International University of Rabat\\
	\normalsize\textit{ilias.tougui@uir.ac.ma}}
 {Mehdi Zakroum}
	{International University of Rabat\\
	\normalsize\textit{mehdi.zakroum@uir.ac.ma}}
  {Mounir Ghogho}
	{Mohammed VI Polytechnic University\\
	\normalsize\textit{mounir.ghogho@um6p.ma}}

\begin{document}

%
\maketitle
\begin{abstract}
Parkinson's Disease (PD) affects over 10 million people worldwide, with speech impairments in up to 89\% of patients. Current speech-based detection systems analyze entire utterances, potentially overlooking the diagnostic value of specific phonetic elements. We developed a granularity-aware approach for multilingual PD detection using an automated pipeline that extracts time-aligned phonemes, syllables, and words from recordings. Using Italian, Spanish, and English datasets, we implemented a bidirectional LSTM with multi-head attention to compare diagnostic performance across the different granularity levels. Phoneme-level analysis achieved superior performance with AUROC of 93.78\% ± 2.34\% and accuracy of 92.17\% ± 2.43\%. This demonstrates enhanced diagnostic capability for cross-linguistic PD detection. Importantly, attention analysis revealed that the most informative speech features align with those used in established clinical protocols: sustained vowels (/a/, /e/, /o/, /i/) at phoneme level, diadochokinetic syllables (/ta/, /pa/, /la/, /ka/) at syllable level, and /pataka/ sequences at word level. Source code will be available at \url{https://github.com/jetliqs/clearpd}\footnote{Source code and model weights to be published with proceedings}.

\end{abstract}
\begin{keywords}
Parkinson's Disease, Speech Analysis, Multi-granularity, Cross-lingual, Interpretability
\end{keywords}

\section{Introduction}
\label{sec:intro}
Parkinson's Disease (PD) is a neurodegenerative disorder that affects more than 10 million people worldwide, with up to 89\% of patients experiencing speech and communication impairments, often preceding motor symptoms for several years \cite{ho1999speech}. Early and accurate diagnosis of PD through speech analysis has emerged as a promising noninvasive approach, with recent studies achieving classification accuracies exceeding 90\% \cite{he2024exploiting}. However, current methodologies predominantly analyze entire speech utterances as single units, overlooking the diagnostic value of specific phonetic elements \cite{favaro2023multilingual, lim2025cross}.

The human speech production system is inherently hierarchical, comprising multiple granularity levels from phonemes to syllables, words, and complete utterances \cite{henry2004verbal, shao2014verbal}. Emerging evidence suggests that speech deterioration related to PD does not affect all phonetic elements uniformly \cite{klumpp2022phonetic}. Certain phonemes, particularly fricatives and plosives, demonstrate greater sensitivity to the motor control deficits characteristic of PD, while others remain relatively preserved in early stages of the disease \cite{moro2019phonetic}. This differential impact may imply that targeted analysis of specific sound combinations may yield better diagnostic performance compared to whole-utterance approaches.

Despite this compelling hypothesis, the vast majority of existing PD speech detection systems employ deep learning (DL) models trained on complete speech samples \cite{he2024exploiting, klumpp2022phonetic}. Only a limited body of research has investigated the diagnostic potential of fine-grained speech granularities, with a study showing that phoneme, syllable and word-level features can achieve classification accuracies up to 86\%, 80\% and 83\% respectively \cite{gallo2024automatic}. This research gap stems primarily from the acute scarcity of datasets labeled at phoneme, syllable, and word levels – a fundamental barrier preventing systematic investigation of granularity-based PD detection approaches.

Furthermore, most existing studies focus on monolingual datasets, limiting the generalizability of findings across diverse linguistic populations. Recent multilingual PD detection research shows that cross-linguistic approaches achieve superior diagnostic performance compared to language-specific models. For instance, a combined Korean-Taiwanese approach \cite{lim2025cross} achieved AUROC of 90\% compared to individual language performance of 87\% - 88\%, while multilingual pretrained models significantly outperformed monolingual variants in early PD detection. These findings suggest that certain speech biomarkers associated with motor impairment in PD may be language independent.

\begin{figure*}[ht]
  \centering
  \includegraphics[width=\textwidth]{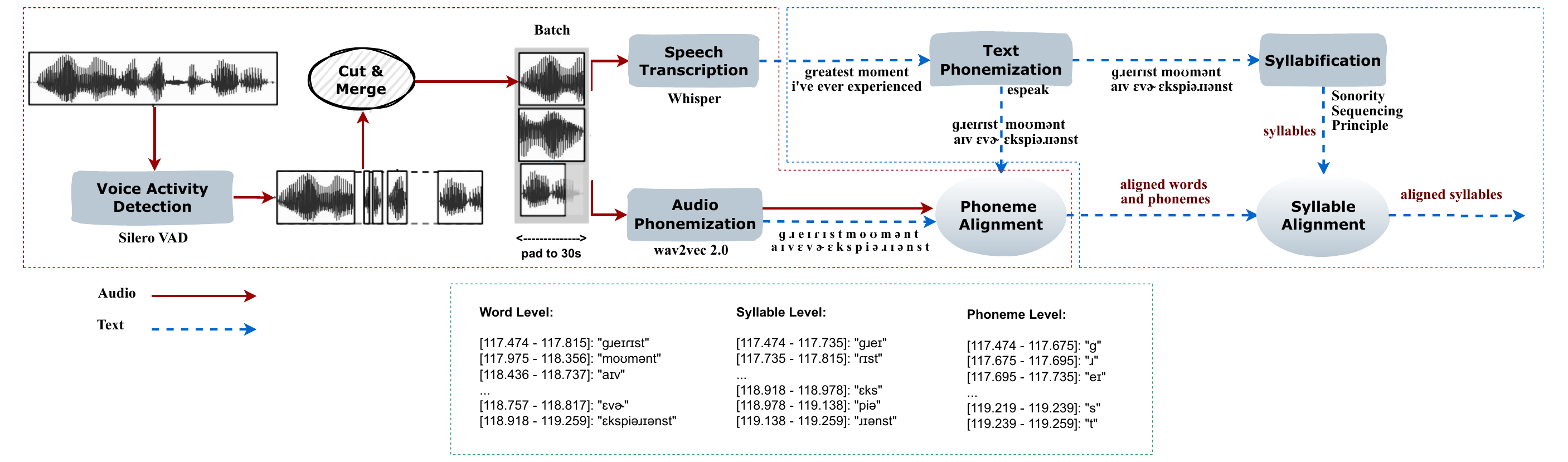}
  \vspace{-7mm}
  \caption{Speech units extraction framework: Recordings are processed through voice activity detection, transcription, phonemization and syllabification to infer multi-granular speech units like words, syllables and phonemes with temporal boundaries.}
  \label{fig:pipeline}
\end{figure*}

This paper addresses these limitations by introducing a granularity-aware approach for PD speech detection. We present an automated pipeline that extracts time-aligned phonemes, syllables, and words from speech recordings, enabling systematic comparison of diagnostic performance across multiple granularity levels. Leveraging publicly available datasets in Italian \cite{italianpd2019}, Spanish \cite{mendes2024neurovoz}, and English \cite{jaeger2019mobile}, our framework facilitates cross-linguistic investigation of how granularity exhibits PD speech biomarkers. The key contributions are: (1) an automatic pipeline for generating time-aligned multi-granularity speech annotations, (2) the first systematic comparison of multi-granularity speech features for multilingual PD detection, and (3) preliminary evidence that specific granularities offer promising diagnostic potential that aligns with established clinical practices in PD diagnosis.

\vspace{-1mm}
\section{Recognizing Parkinson's from Speech}
To enable systematic analysis of PD speech at multiple granularity levels, we developed a modular pipeline that processes raw audio recordings into time-aligned words, syllables, and phonemes, as illustrated in Fig.~\ref{fig:pipeline}. The audio transcription stage was inspired by WhisperX~\cite{bain2023whisperx}. Our language-agnostic framework integrates state-of-the-art components within a modular design that enables efficient processing of speech data while maintaining temporal alignment across all speech units. The key components are described in the following.


\textbf{Voice Activity Detection}. Audio recordings are processed using Silero VAD\footnote{https://github.com/snakers4/silero-vad}, a pre-trained model employing a hybrid architecture that detects speech segments in 16 kHz waveforms. The model processes audio in 512-sample windows (32 ms) and outputs a probability \(P_s\) for speech presence, using a threshold of \(0.5\). Segments exceeding 30 seconds are split, while shorter segments are merged up to this limit for later batch processing.

\textbf{Automatic Speech Transcription}. Segmented audio batches are transcribed using Whisper \cite{whisper}, an encoder-decoder Transformer trained on 680,000 hours of weakly-supervised multilingual speech data. The model processes 30-second audio chunks by converting them to 80-channel log-mel spectrograms, which are then encoded and decoded to produce word-level aligned transcriptions. We employed the Whisper large-v3 model, achieving Word Error Rates (WER) of 4.7\% for Spanish, 5.5\% for Italian, and 9.3\% for English on the CommonVoice dataset \cite{whisper}.

\textbf{Audio and Text Phonemization}. To obtain phoneme-level alignment, we employed the wav2vec 2.0 framework~\cite{baevski2020wav2vec}. This model processes audio with a CNN-based feature encoder followed by a Transformer context network, producing frame-level phoneme probabilities aligned to the input waveform~\cite{baevski2020wav2vec}. The model\footnote{https://huggingface.co/facebook/wav2vec2-lv-60-espeak-cv-ft} was fine-tuned on the CommonVoice dataset with eSpeak\footnote{https://github.com/espeak-ng/espeak-ng} phonemization \cite{Bernard2021}, enabling it to output phonetic sequences in International Phonetic Alphabet (IPA) format with high accuracy over 100 languages~\cite{baevski2020wav2vec}.

\textbf{Syllabification}. Syllable boundaries are assigned using an SSP-based (Sonority Sequencing Principle) syllabification module. The SSP algorithm \cite{selkirk1984major} identifies syllable nuclei (vowels as sonority peaks) and partitions words into constituent syllables by decreasing sonority toward word edges. This approach provides consistent rule-based syllable segmentation across languages.

\textbf{Phoneme and Syllable Alignments}: Temporal alignment of words, phonemes, and syllables across the pipeline is achieved through a combination of CTC (Connectionist Temporal Classification) alignment methods for phonemes and words, and custom rule-based SSP alignment for syllables. This produces synchronized boundaries (starting and ending timestamps) enabling multi-granularity analysis at the word, syllable, and phoneme levels.

\textbf{The Prediction Model}. We implemented a bidirectional LSTM with multi-head attention for granularity-based PD detection. The model consists of a 6-layer bidirectional LSTM with 512 hidden units and 0.3 dropout for regularization. To handle variable-length sequences efficiently, we employed packed sequence processing, which eliminates computational overhead from padding tokens. Following the LSTM layers, an 8-head attention mechanism performs sequence-level feature aggregation, enabling the model to focus on discriminative speech patterns within each sequence.

\begin{figure}[t]
  \centering
  \includegraphics[scale=0.55]{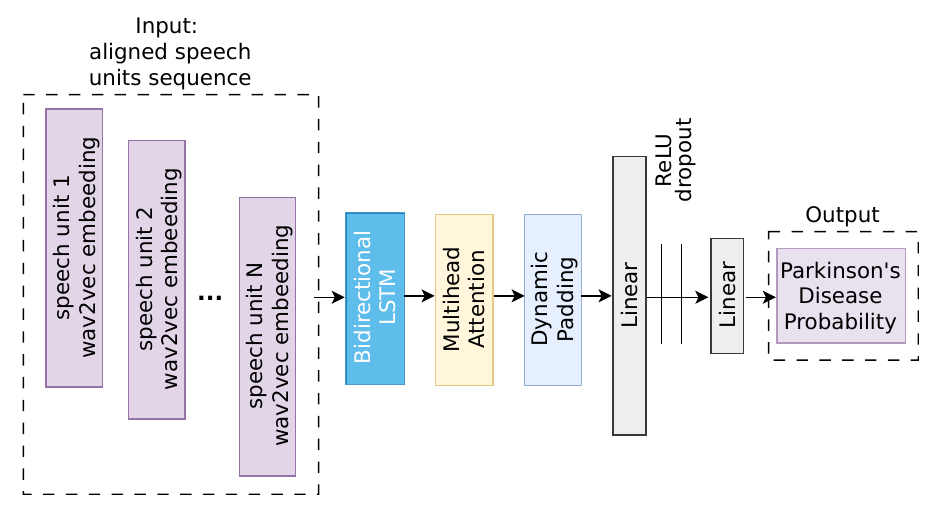}
  \vspace{-8mm}
  \caption{Architecture of the Parkinson's Disease Prediction Model: a bidirectional LSTM with multi-head attention}
  \label{fig:model}
\end{figure}

\vspace{-3mm}
\section{Experimental Setup}
In the following, we present the experimental setup, highlighting the key data preprocessing steps, the hyperparameter tuning methodology, and the training and evaluation procedures used to assess the performance of our model.

\textbf{Datasets}. We employ three publicly available multilingual PD speech datasets comprising recordings in Italian \cite{italianpd2019}, Spanish \cite{mendes2024neurovoz}, and English \cite{jaeger2019mobile}. Each dataset contains speech samples from both PD patients and healthy controls (HC). The datasets exhibit natural variability in recording duration and speech content (scripted reading passages, spontaneous dialogue and open monologue), providing a robust foundation for cross-linguistic analysis.

\textbf{Sequence Preprocessing}. Following granularity extraction, we implemented a feature extraction pipeline using the XLSR-53 model\footnote{https://huggingface.co/facebook/wav2vec2-large-xlsr-53}, a cross-lingual variant of wav2vec 2.0 trained on 53 languages for multilingual feature extraction. We extracted features from multiple transformer layers [0, 6, 12, 18, 24], with the optimal layer selected via hyperparameter tuning on the validation set. To ensure data quality, we applied confidence-based filtering with a threshold of 0.5, retaining only phonemes with reliable alignment scores.


\textbf{Data Splitting Strategy}. We implemented a stratified speaker-independent splitting strategy to prevent data leakage and ensure robust evaluation. The splitting procedure grouped recordings by speaker identity and created balanced partitions across multiple stratification factors: diagnosis label (PD/HC), language, and recording duration bins. We used a split of 60\% training, 20\% validation, and 20\% test, with no speaker appearing in multiple splits. Variable-length sequences were handled through dynamic padding during batch creation, with attention masks preserving the original sequence boundaries for model training.

\textbf{Training Configuration}. The model was trained using AdamW optimizer with L2 weight decay (0.01) and a learning rate of 1e-5. We applied ReduceLROnPlateau scheduling with factor=0.5 and patience=5 to adapt the learning rate based on validation loss plateaus. Training employed early stopping based on validation F1-score to prevent overfitting. Gradient clipping (max norm=1.0) was applied to stabilize training and prevent gradient explosion. The model was trained with batch size 32 for up to 15 epochs, using cross-entropy loss for binary classification.

\textbf{Hyperparameter Optimization}. LSTM-specific hyperparameters were optimized through systematic validation on held-out data. We used XLSR-53 layer 12 representations as input features (1024-dimensional), selected based on preliminary experiments showing optimal performance at this depth. The confidence threshold for segment filtering was set to 0.6, balancing data quality with sample retention.

\textbf{Evaluation Protocol}. Model performance was assessed using subject-level aggregation, where predictions from multiple speech segments per speaker were averaged before final classification to ensure clinical relevance by simulating real-world diagnostic scenarios where multiple speech samples inform patient-level decisions. We employed speaker-independent data splits to prevent information leakage and report comprehensive metrics including Accuracy, F1-score, AUROC, and AUPRC computed at the subject level. Attention weights were extracted during inference to enable interpretability analysis of which speech segments contributed most to PD detection decisions.

\vspace{-1mm}
\section{Results}
We evaluate the model performance across different speect units --phoneme, syllable, and word-- using speaker-independent test sets. Each configuration is trained five times with different random seeds, and results are reported as mean ± standard deviation.

\begin{table}[!htb]
\centering
\label{tab:auroc_auprc}
\caption{Model Performance - AUROC and AUPRC}
\vspace{2 mm}
\resizebox{\columnwidth}{!}{
\begin{tabular}{lcc}
\toprule
\textbf{Granularity} & \textbf{AUROC} & \textbf{AUPRC} \\
\midrule
Phoneme & \textbf{0.9378 $\pm$ 0.0234} & 0.9404 $\pm$ 0.0337 \\
Syllable & 0.9212 $\pm$ 0.0172 & \textbf{0.9455 $\pm$ 0.0135} \\
Word & 0.9222 $\pm$ 0.0066 & 0.9364 $\pm$ 0.0129 \\
\bottomrule
\end{tabular}
}
\end{table}
\vspace{-7mm}
\begin{table}[!htb]
\centering
\label{tab:f1_acc}
\caption{Model Performance - F1 and ACC}
\vspace{2mm}
\resizebox{\columnwidth}{!}{
\begin{tabular}{lcc}
\toprule
\textbf{Granularity} & \textbf{F1} & \textbf{ACC} \\
\midrule
Phoneme & \textbf{0.9213 $\pm$ 0.0249} & \textbf{0.9217 $\pm$ 0.0243} \\
Syllable & 0.9074 $\pm$ 0.0287 & 0.9079 $\pm$ 0.0284 \\
Word & 0.8873 $\pm$ 0.0170 & 0.8875 $\pm$ 0.0171 \\
\bottomrule
\end{tabular}
}
\end{table}

Tables 1 and 2 present the comprehensive evaluation results. Phoneme-level analysis achieved the highest discriminative performance with AUROC of 93.78\% ± 2.34\% and accuracy of 92.17\% ± 2.43\%, demonstrating superior capability in capturing PD-related speech patterns. Syllable-level granularity obtained the highest AUPRC (94.55\% ± 1.35\%) while maintaining competitive performance across other metrics. Word-level analysis showed the most conservative results with the lowest F1-score (88.73\% ± 1.70\%) and accuracy (88.75\% ± 1.71\%).

The low standard deviations across all metrics (ranging from 0.66\% to 3.37\%) confirm the statistical reliability and reproducibility of our approach across different data splits. All granularity levels exceeded 88\% accuracy, indicating clinically relevant performance for automated PD screening applications.

The superior performance of phoneme-level features validates our hypothesis that fine-grained speech analysis provides enhanced diagnostic capability. The AUROC values exceeding 92\% for phoneme and syllable levels suggest strong potential for real-world deployment in clinical settings.

\vspace{-1mm}
\section{Discussion}
\begin{figure}[ht]
    \centering
    \includegraphics[width=\columnwidth]{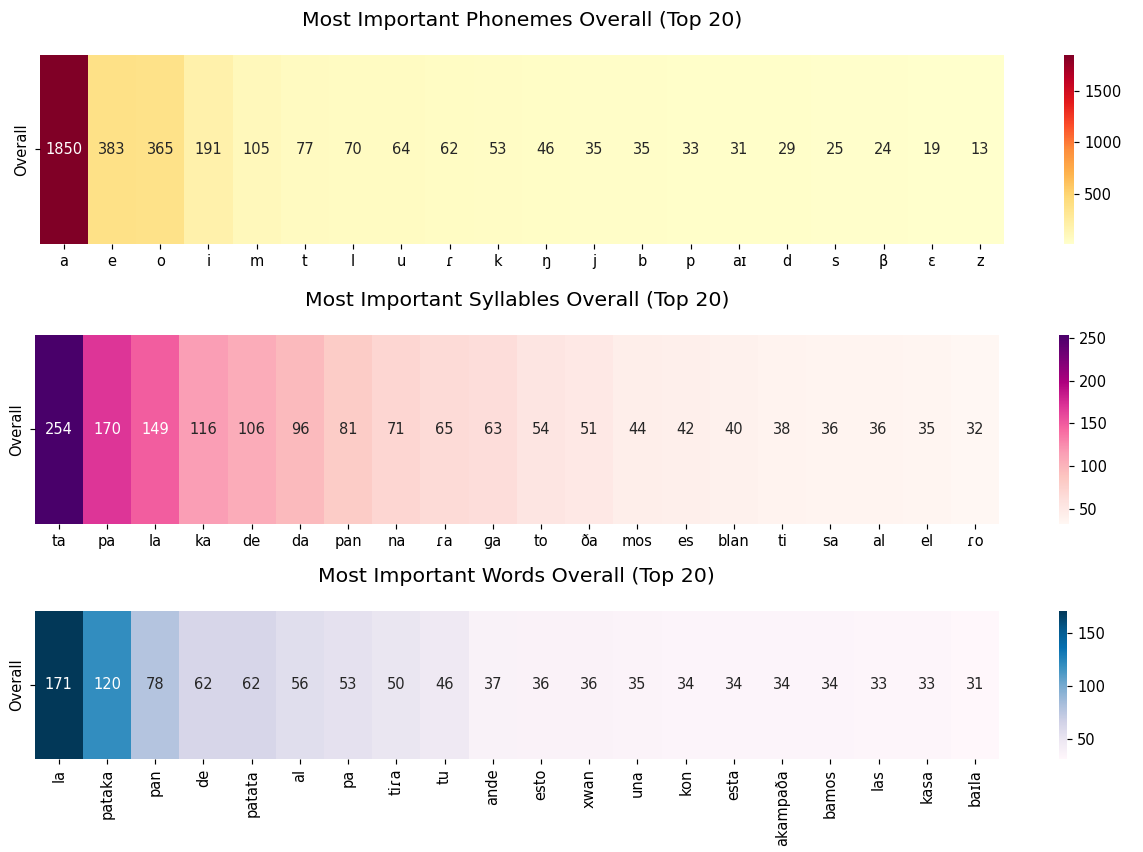}
    \vspace{-10mm}
    \caption{Multi-granularity attention weights of the model on the test set for PD recognition. Heat maps show importance rankings of top 20 phonemes, syllables, and words, with color intensity indicating attention scores.}
    \label{fig:attention}
\end{figure}

As shown in Fig.\ref{fig:attention}, our multi-granular cross-lingual attention mechanism successfully identified diagnostically relevant speech features that align remarkably with established clinical practices in PD diagnosis. Critically, these findings emerge from a multilingual dataset combining English, Italian, and Spanish. The convergence between our AI-based data-driven approach and decades of clinical research validates both our methodology and existing diagnostic protocols, providing an automatic framework for assisting experts in inferring PD from speech.

At the phoneme level, the model prioritized sustained vowels \textbf{/a/} (1850), \textbf{/e/} (383), \textbf{/o/} (365), and \textbf{/i/} (191) across different linguistic contexts, directly supporting clinical literature on sustained phonation tasks \cite{brzoskowski2025speech}. These vowels effectively reveal core phonatory impairments—reduced vocal cord vibration, breathiness, and altered pitch variability. The highest attention weight given to \textbf{/a/} reflects its widespread use in clinical protocols, where it serves as the primary vowel for voice quality assessment due to its optimal acoustic properties for detecting subtle changes in vocal fold function and respiratory control. The consonant phonemes \textbf{/i/}, \textbf{/m/}, \textbf{/t/}, \textbf{/l/}, \textbf{/f/},\textbf{/k/} in the middle-to-lower importance range align with research showing that imprecise consonants are a key hallmark of PD speech, with fricatives \textbf{/f/} and plosives (\textbf{/t/}, \textbf{k}) being particularly sensitive to motor impairments. Their moderate ranking indicates meaningful diagnostic contribution while being less dominant than vowels \cite{cao2025speech}.

At the syllable level, highest attention weights were assigned to \textbf{/ta/} (254), \textbf{/pa/} (170), \textbf{/la/} (149), and \textbf{/ka/} (116) across the dataset, corresponding precisely to diadochokinetic (DDK) task components used in clinical practice \cite{brzoskowski2025speech}. These syllables challenge articulatory precision and motor coordination, capturing disease-specific deficits in speech timing and accuracy. The model's focus on these specific syllables demonstrates its ability to identify the fundamental motor speech patterns that clinicians rely upon for assessing articulatory agility, tongue-tip coordination, and rapid movement sequencing—all key indicators of neuromotor decline in PD. The syllables \textbf{/de/}, \textbf{/da/}, \textbf{/pan/}, \textbf{/na/}, \textbf{/ra/}, \textbf{/ga/} represent diverse articulatory challenges involving dental/alveolar precision, complex tongue movement, and velar coordination. While less diagnostically powerful than DDK combinations, they still capture meaningful articulatory deficits characteristic of PD motor impairment \cite{cao2025speech}.

Even though the training corpus spanned a wide spectrum of speech tasks \textit{(scripted reading passages, spontaneous dialogue and open monologue)}, and despite competing with far longer and linguistically richer material, the \textbf{/pataka/} (171) sequence still attracted the greatest attention. At the word level, the model consistently singled out this rapid diadochokinetic token for its ability to probe the full articulatory range—bilabial \textbf{/pa/}, alveolar \textbf{/ta/} and velar \textbf{/ka/}—within a single breath. In other words, the model’s focus on \textbf{/pataka/} is not a sampling artifact but a data-driven confirmation that this compact exercise remains the most efficient acoustic proxy for global oral-motor control in PD assessment, which directly confirms the literature \cite{brzoskowski2025speech}.

\vspace{-3mm}
\section{Conclusion}
In this study, we designed and developed a multilingual approach for detecting PD that considers varying levels of granularity, utilizing an automated process to extract phonemes, syllables, and words aligned with audio recordings.
While our results demonstrate strong alignment with clinical literature across three major languages, future work should expand linguistic coverage to include low-resource languages where PD diagnostic tools are critically needed. In addition, clinical validation and integration with existing assessment tools (MDS-UPDRS), and extension to differential diagnosis capabilities distinguishing PD from other movement disorders represent essential next steps.


\bibliographystyle{IEEEbib}
\bibliography{strings,refs}

\begin{thebibliography}{10}

\bibitem{ho1999speech}
Aileen~K Ho, Robert Iansek, Caterina Marigliani, John~L Bradshaw, and Sandra Gates,
\newblock ``Speech impairment in a large sample of patients with parkinson’s disease,''
\newblock {\em Behavioural neurology}, vol. 11, no. 3, pp. 131--137, 1999.

\bibitem{he2024exploiting}
Tongyue He, Junxin Chen, Xu~Xu, and Wei Wang,
\newblock ``Exploiting smartphone voice recording as a digital biomarker for parkinson’s disease diagnosis,''
\newblock {\em IEEE Transactions on Instrumentation and Measurement}, vol. 73, pp. 1--12, 2024.

\bibitem{favaro2023multilingual}
Anna Favaro, Laureano Moro-Vel{\'a}zquez, Ankur Butala, Chelsie Motley, Tianyu Cao, Robert~David Stevens, Jes{\'u}s Villalba, and Najim Dehak,
\newblock ``Multilingual evaluation of interpretable biomarkers to represent language and speech patterns in parkinson's disease,''
\newblock {\em Frontiers in Neurology}, vol. 14, pp. 1142642, 2023.

\bibitem{lim2025cross}
Wee~Shin Lim, Shu-I Chiu, Pei-Ling Peng, Jyh-Shing~Roger Jang, Sol-Hee Lee, Chin-Hsien Lin, and Han-Joon Kim,
\newblock ``A cross-language speech model for detection of parkinson’s disease,''
\newblock {\em Journal of Neural Transmission}, vol. 132, no. 4, pp. 579--590, 2025.

\bibitem{henry2004verbal}
Julie~D Henry and John~R Crawford,
\newblock ``Verbal fluency deficits in parkinson's disease: a meta-analysis,''
\newblock {\em Journal of the International Neuropsychological Society}, vol. 10, no. 4, pp. 608--622, 2004.

\bibitem{shao2014verbal}
Zeshu Shao, Esther Janse, Karina Visser, and Antje~S Meyer,
\newblock ``What do verbal fluency tasks measure? predictors of verbal fluency performance in older adults,''
\newblock {\em Frontiers in psychology}, vol. 5, pp. 772, 2014.

\bibitem{klumpp2022phonetic}
Philipp Klumpp, Tom{\'a}s Arias-Vergara, Juan~Camilo V{\'a}squez-Correa, Paula~Andrea P{\'e}rez-Toro, Juan~Rafael Orozco-Arroyave, Anton Batliner, and Elmar N{\"o}th,
\newblock ``The phonetic footprint of parkinson’s disease,''
\newblock {\em Computer Speech \& Language}, vol. 72, pp. 101321, 2022.

\bibitem{moro2019phonetic}
Laureano Moro-Velazquez, Jorge~A Gomez-Garcia, Juan~I Godino-Llorente, Francisco Grandas-Perez, Stefanie Shattuck-Hufnagel, Virginia Yag{\"u}e-Jimenez, and Najim Dehak,
\newblock ``Phonetic relevance and phonemic grouping of speech in the automatic detection of parkinson’s disease,''
\newblock {\em Scientific reports}, 2019.

\bibitem{gallo2024automatic}
Jeferson~David Gallo-Aristiz{\'a}bal, Daniel Escobar-Grisales, Cristian~David R{\'\i}os-Urrego, Elmar N{\"o}th, and Juan~Rafael Orozco-Arroyave,
\newblock ``Automatic classification of parkinson’s disease using wav2vec embeddings at phoneme, syllable, and word levels,''
\newblock in {\em International Conference on Text, Speech, and Dialogue}. Springer, 2024, pp. 313--323.

\bibitem{italianpd2019}
Giovanni Dimauro and Francesco Girardi,
\newblock ``Italian parkinson's voice and speech,'' 2019.

\bibitem{mendes2024neurovoz}
Jana{\'\i}na Mendes-Laureano, Jorge~A G{\'o}mez-Garc{\'\i}a, Alejandro Guerrero-L{\'o}pez, Elisa Luque-Buzo, Juli{\'a}n~D Arias-Londo{\~n}o, Francisco~J Grandas-P{\'e}rez, and Juan~I Godino-Llorente,
\newblock ``Neurovoz: a castillian spanish corpus of parkinsonian speech,''
\newblock {\em Scientific Data}, vol. 11, no. 1, pp. 1367, 2024.

\bibitem{jaeger2019mobile}
Hagen Jaeger, Dhaval Trivedi, and Michael Stadtschnitzer,
\newblock ``Mobile device voice recordings at king’s college london (mdvr-kcl) from both early and advanced parkinson’s disease patients and healthy controls,''
\newblock {\em Zenodo}, 2019.

\bibitem{bain2023whisperx}
Max Bain, Jaesung Huh, Tengda Han, and Andrew Zisserman,
\newblock ``Whisperx: Time-accurate speech transcription of long-form audio,''
\newblock {\em arXiv preprint arXiv:2303.00747}, 2023.

\bibitem{whisper}
Alec Radford, Jong~Wook Kim, Tao Xu, Greg Brockman, Christine McLeavey, and Ilya Sutskever,
\newblock ``Robust speech recognition via large-scale weak supervision,''
\newblock in {\em International conference on machine learning}. PMLR, 2023, pp. 28492--28518.

\bibitem{baevski2020wav2vec}
Alexei Baevski, Yuhao Zhou, Abdelrahman Mohamed, and Michael Auli,
\newblock ``wav2vec 2.0: A framework for self-supervised learning of speech representations,''
\newblock {\em Advances in neural information processing systems}, vol. 33, pp. 12449--12460, 2020.

\bibitem{Bernard2021}
Mathieu Bernard and Hadrien Titeux,
\newblock ``Phonemizer: Text to phones transcription for multiple languages in python,''
\newblock {\em Journal of Open Source Software}, vol. 6, no. 68, pp. 3958, 2021.

\bibitem{selkirk1984major}
Elisabeth Selkirk,
\newblock ``On the major class features and syllable theory,''
\newblock {\em Language sound structure}, 1984.

\bibitem{brzoskowski2025speech}
Vanessa Brzoskowski~dos Santos, Amanda~Lara Bressanelli, Fernanda~Venzke Zardin, Rui Rothe-Neves, and Maira~Rozenfeld Olchik,
\newblock ``Speech characteristics across motor subtypes of parkinson's disease,''
\newblock {\em International journal of language \& communication disorders}, vol. 60, no. 4, pp. e70081, 2025.

\bibitem{cao2025speech}
Fangyuan Cao, Adam~P Vogel, Puya Gharahkhani, and Miguel~E Renteria,
\newblock ``Speech and language biomarkers for parkinson’s disease prediction, early diagnosis and progression,''
\newblock {\em npj Parkinson's Disease}, vol. 11, no. 1, pp. 57, 2025.

\end{thebibliography}
\label{sec:refs}

\end{document}